\documentclass[11pt,a4paper, dvipsnames]{article}
\usepackage{times,latexsym,paralist}
\usepackage{url}
\usepackage[T1]{fontenc}

\usepackage[acceptedWithA]{tacl2021v1}
\usepackage{xspace,mfirstuc,tabulary}
\usepackage{inconsolata}
\usepackage{xcolor}
\usepackage{graphicx}
\usepackage{multirow}
\usepackage{booktabs}
\usepackage{amsmath,amsfonts,amssymb}
\usepackage{hyperref}

\newif\iftaclinstructions
\taclinstructionsfalse % AUTHORS: do NOT set this to true
\iftaclinstructions
\renewcommand{\confidential}{}
\renewcommand{\anonsubtext}{(No author info supplied here, for consistency with
TACL-submission anonymization requirements)}
\newcommand{\instr}
\fi

\iftaclpubformat % this "if" is set by the choice of options

\else

\fi

\definecolor{crimson}{rgb}{0.86, 0.08, 0.24}

\title{Exploring Contrast Consistency of Open-Domain Question Answering Systems on Minimally Edited Questions}
\author{{\bf Zhihan Zhang, Wenhao Yu, Zheng Ning, Mingxuan Ju, Meng Jiang} \\
Department of Computer Science and Engineering, University of Notre Dame \\
\normalsize{ {\tt \{zzhang23, wyu1, zning, mju2, mjiang2\}@nd.edu}}
}

\date{}

\begin{document}
\maketitle
\begin{abstract}
Contrast consistency, the ability of a model to make consistently correct predictions in the presence of perturbations, is an essential aspect in NLP. While studied in tasks such as sentiment analysis and reading comprehension, it remains unexplored in open-domain question answering (OpenQA) due to the difficulty of collecting perturbed questions that satisfy factuality requirements. In this work, we collect minimally edited questions as challenging contrast sets to evaluate OpenQA models. Our collection approach combines both human annotation and large language model generation. We find that the widely used dense passage retriever (DPR) performs poorly on our contrast sets, despite fitting the training set well and performing competitively on standard test sets. To address this issue, we introduce a simple and effective query-side contrastive loss with the aid of data augmentation to improve DPR training. Our experiments on the contrast sets demonstrate that DPR's contrast consistency is improved without sacrificing its accuracy on the standard test sets.\footnote{Awaiting publication at TACL. This is a pre-MIT Press publication version. Data and code are available at \url{https://github.com/ytyz1307zzh/Minimally_Edited_Questions}.}

% TACL关于数据、代码的要求：To keep reviewing times down, we do not allow submission of supplementary material. (Furthermore, we disallow links to such supplementary material, anonymized or not, to avoid the possibility of tracking analytics compromising the anonymity of the reviewers.) However,  the release of software or datasets is strongly encouraged.  We ask that authors include in their submission some text that explains, in anonymous terms, whether the software or data will be released, and if so, in what fashion.  The review forms will prompt the referees to take this into account. 所以我无法给出匿名链接。
\end{abstract}

\section{Introduction}
\label{sec:introduction}
\begin{figure}[ht!]
    \centering
    \includegraphics[width=0.5\textwidth]{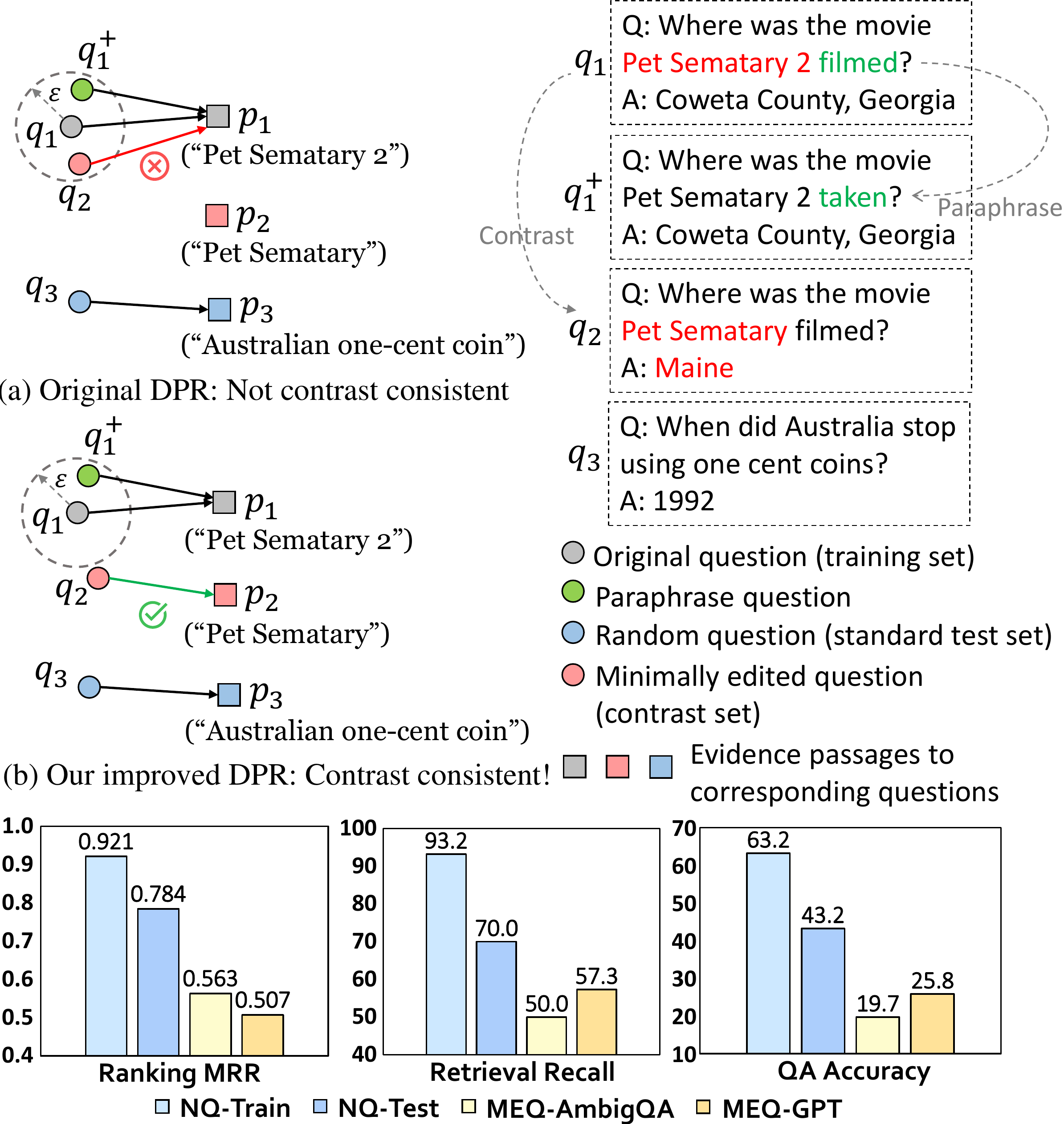}
    \caption{Above: Trained on question $q_1$ but not a contrast one $q_2$, DPR generated an overly similar embedding of $q_2$ with $q_1$'s and thus falsely retrieved $p_1$. We aim to identify $q_2$ as a distinct question and retrieve $p_2$ instead. Below: The performance of DPR-based OpenQA models on the standard NQ question set and our contrast set of minimally edited questions (MEQs).}
    \label{fig:example}
\end{figure}

Contrast consistency~\cite{contrast_data_decision_boundary} is a crucial aspect for neural models in NLP. Models are expected to identify perturbations in the text input and decide whether such a semantic shift leads to a different label. To evaluate this consistency, contrast sets have been introduced in various tasks such as sentiment analysis~\cite{PolyJuice}, natural language inference~\cite{Tailor}, and reading comprehension~\cite{usa_attack_taiwan} by minimally modifying the original input to reverse the original label. However, to our best knowledge, there is no study on the contrast consistency in open-domain question answering (OpenQA). In OpenQA, even a slight modification of a word or two can alter the meaning of the question, which leads to a completely different answer. To maintain contrast consistency, models are expected to predict the corresponding answer when such semantic shift occurs.

Studying contrast consistency in OpenQA poses unique challenges. Firstly, collecting appropriate contrast sets is difficult. While contrast sets have been developed for reading comprehension~\cite{usa_attack_taiwan, LLM_ControllableQA}, they typically replaced an entity (\textit{e.g.}, Barack Obama was born in \textit{Hawaii}) in given context with another entity (\textit{e.g.}, Barack Obama was born in \textit{New York}), leading to a different answer to the given question (\textit{e.g.}, Where was Barack Obama born?). Constructing such contrast sets does not necessitate the factuality of the perturbed context, as the answer depends solely on the context rather than world knowledge. However, in the absence of evidence context, the perturbed questions in OpenQA must be factually answerable in accordance with world knowledge, which is beyond what rule-based methods can do. Secondly, achieving contrast consistency is challenging for OpenQA models which usually follow the ``retrieve-then-read'' pipeline~\cite{RAG}. In addition to the challenge of predicting answers from a contrast context as in reading comprehension, models also face the challenge of mapping the perturbed question with its corresponding evidence passage in a large corpus. The latter requires the retriever to distinguish the minimal semantic difference between embeddings of the perturbed question and the original question, which is ignored in typical retriever training.

% Compared to related fields like reading comprehension, maintaining contrast consistency in OpenQA presents a more formidable challenge. Firstly, OpenQA questions are not grounded in a specific context. Unlike their reading comprehension counterparts~\cite{usa_attack_taiwan, LLM_ControllableQA}, OpenQA models do not have access to a context that is modified to contain the new answer. Secondly, OpenQA models typically use dense retrievers which compress a question into a single vector to find evidence text~\cite{DPR, contriever}. This makes the retriever insensitive to minimal edits in the question which does not cause much semantic shift but changes the answer. Additionally, collecting contrasting data presents a further challenge in the examination of contrast consistency in OpenQA. Unlike reading comprehension where synthetic evidence can be generated by manipulating the context~\cite{usa_attack_taiwan}, the perturbed questions in OpenQA must still be factually reasonable in accordance with world knowledge.
To fill this gap in OpenQA, we propose to create contrast sets using \textbf{M}inimally \textbf{E}dited \textbf{Q}uestions (MEQs). Given a question $q$ and its answer $a$, an MEQ $q'$ is defined as a question that possesses high lexical and semantic similarity with $q$, while having a distinct answer $a'$ ($a'\neq a$). For example, in Figure~\ref{fig:example}, changing ``Pet Sematary 2'' to ``Pet Sematary'' generates an MEQ that resembles the original question but has a distinct answer (``Coweta County, Georgia''$\rightarrow$``Maine''). We use the training set of an existing benchmark as the original questions because neural OpenQA models exhibit high performance on them. Thus, we are able to evaluate the models' ability of distinguishing MEQs by measuring their performance on the MEQ contrast set. Specifically, we collect MEQs for training questions in the Natural Questions (NQ) benchmark~\cite{NaturalQuestions} from two sources, namely (1) InstructGPT-based question generation~\cite{instructgpt} then crowdsource annotation and (2) the AmbigQA dataset~\cite{AmbigQA}.

We find that the state-of-the-art OpenQA models which employ the dense passage retriever (DPR)~\cite{DPR} struggle on our MEQ contrast sets. As shown in Figure~\ref{fig:example}, DPR-retrieved passages lead to 63\% downstream QA accuracy on training set and 43\% on standard test set. However, the accuracy drops to 20\%\textasciitilde25\% on our MEQ contrast sets.
The problem lies in the contrastive training process of DPR. The model is trained to optimize question embeddings to be closer to their positive passage embeddings\footnote{A passage that provides evidence for answering the question is its positive passage, otherwise a negative passage.} than negative passage embeddings. This paradigm does not provide explicit signals for understanding the relationships between questions, which causes the generated question embeddings to be insensitive to minimal discrepancies. As a result, the model generates overly similar embeddings for the MEQ and the original question, leading to incorrect passage retrieval for the MEQ. In fact, the overlap between the retrieved passages of the original question and those of its MEQ is as high as \textasciitilde70\%, which reflects DPR's limited ability in distinguishing the questions. To overcome such limitations, it is necessary to complement DPR training with signals on inter-question relationships. Besides building the mapping between questions and passages, DPR needs to know which questions are the same and which are different.

In this pioneering study, we propose a simple and effective method based on a query-side contrastive loss to improve the performance of DPR on MEQs. Specifically, in order to learn inter-question relationships, DPR is trained to distinguish between paraphrase questions and semantically different questions. To achieve this, we obtain synthetic MEQs for training questions from the machine-created QA corpus, PAQ~\cite{PAQ}, as augmented data. Experiments demonstrate that learning the query-side contrastive loss on the augmented MEQs improves the performance of DPR on contrast sets, without sacrificing its performance on standard open-domain questions in the NQ test set.

\section{Related Work}
% Background for DPR (bi-encoder architecture and the contrastive loss)

\subsection{Open-Domain Question Answering}

OpenQA is a task that aims to answer user questions without any specified context, thereby testing the ability of QA systems to retrieve, comprehend, and utilize world knowledge~\cite{OpenQA_review}. The state-of-the-art approach in OpenQA is a two-stage pipeline, consisting of \textit{evidence retrieval} and \textit{answer prediction}~\cite{DrQA}. 

In the evidence retrieval stage, a retriever model finds evidence passages from a large corpus (\textit{e.g.}, Wikipedia) based on their relevance to the question. Traditional retrievers like BM25~\cite{BM25} perform lexical matching to measure such relevance scores. Recently, DPR~\cite{DPR} revolutionized the field by employing dual BERT~\cite{BERT} encoders to compute embeddings for the question and the passage, respectively. It searches evidence passages based on the inner product of question and passage embeddings. Despite subsequent approaches have sought to improve the architecture of the retriever by using fine-grained question-passage interactions~\cite{ColBERT} or enhancing global embedding training~\cite{Condenser}, DPR remains the most widely-used model due to its simplicity and efficiency. 
However, the capability of DPR in distinguishing contrastive information has not been thoroughly studied.
In this work, we use MEQs as contrast sets and show that DPR has limited contrast consistency when solving MEQs.

In the answer prediction stage, a reader model encodes and fuses the representations of all passages, then predicts an answer by extracting a span~\cite{FiE}, generating a free-form sequence~\cite{FiD}, or using a hybrid approach~\cite{R2-D2}. While answer prediction is also challenging on MEQs, our approach mainly focuses on the retrieval part which is the bottleneck of solving the MEQs in OpenQA.

\subsection{Contrast Sets}

NLP Benchmark datasets are typically comprised of \textit{i.i.d.} examples that are randomly divided into training and test sets. Conversely, contrast sets refer to data created from small yet label-changing modifications to the existing examples~\cite{contrast_data_decision_boundary}. 
% Therefore, these examples are usually located close to some original examples within the task data space while having a different label. 
Such characteristics make contrast sets an ideal testbed for evaluating contrast consistency. For example, \citet{contrast_data_decision_boundary} and \citet{counterfactual_revised_data} employed humans to modify linguistic patterns on tasks like syntactic parsing, relation extraction, and claim verification.
On sentiment analysis and language inference tasks, controlled text modification models could automatically generate contrast sets~\cite{PolyJuice, Tailor}. In reading comprehension, rule-based algorithms created contrast sets by replacing the answer with another entity~\cite{usa_attack_taiwan, counterfactual_explain_attribution, LLM_ControllableQA}. In video-to-text matching, a pre-trained T5 model was used to find replacements for verbs and entities in the original caption~\cite{video-text_contrast}.

Nevertheless, building contrast sets to evaluate contrast consistency in OpenQA has not been explored yet, where data collection must guarantee the factuality of MEQs. The most relevant work is~\cite{counterfactual_generation_QA} which automatically generated perturbed questions for data augmentation on QA datasets. However, we focus on collecting challenging MEQs to evaluate model consistency instead of data augmentation.
% which is orthogonal to our motivation. In fact, their approach can serve as an alternative data augmentation method to our simple approach in \S\ref{sec:data_augment}.
Moreover, their generated questions did not meet the requirements of MEQs. The limited accuracy of the question generation model would lead to lots of noise instead of perfect factuality. Also, their method did not ensure the minimality of edits. Therefore, their generated data cannot be used as challenging contrast sets to evaluate contrast consistency in OpenQA.

% Contrast sets are valuable in multiple applications. \textbf{(1) Evaluate model generalization}. Evaluating model performance on contrast sets can reflect whether the model relies on spurious correlations in solving the i.i.d. test set~\cite{counterfactual_revised_data, contrast_data_decision_boundary, video-text_contrast}. \textbf{(2) Explain model predictions}. If a minimal edit causes a model's prediction to change from $p$ to $p'$, the edited words may be used to explain why the model predicts $p$ instead of $p'$~\cite{MICE, counterfactual_explain_attribution}. \textbf{(3) Data Augmentation}. Automatically generated contrast examples, despite their inherent noise, can be used to augment training data which helps improve model generalization and reduce overfitting to spurious correlations~\cite{symmetric_consistency_QA, PolyJuice, Tailor}.

% Obtaining contrast sets is also a challenge since such data are not typically available in public corpora. Although manual annotation requires extensive human labor and careful instruction design to annotators, the resulting data are usually of high quality~\cite{counterfactual_revised_data, contrast_data_decision_boundary}. As an alternative approach, automatic collection of contrast data is also explored by either designing rule-based edits~\cite{symmetric_consistency_QA, usa_attack_taiwan} or training models for controlled text modification~\cite{PolyJuice, Tailor}.

\section{Task: Contrast Consistency on MEQs}

\subsection{Problem Formulation}
\label{sec:formulation}

In this work, we study minimally edited questions (MEQ) as challenging contrast sets in OpenQA. Suppose we have two questions $q$ and $q'$ with answers $a$ and $a'$ respectively, where $q$ is the original question in the training set and $q'$ is an MEQ of $q$. In this study, the minimality of edits is measured in two aspects: lexical distance $d_{\ell}(q, q')$ and semantic distance $d_s(q, q')$. That is to say, $q'$ needs to satisfy $d_{\ell}(q, q') \leq \epsilon_{\ell}$, $d_s(q, q') \leq \epsilon_{s}$ and $a'\neq a$, where $\epsilon_{\ell}$ and $\epsilon_{s}$ are distance thresholds.

\subsection{Evaluation Metrics}
\label{sec: evaluation}

To evaluate DPR on MEQ contrast sets, we consider metrics on both ranking and retrieval evaluation. Besides, we run end-to-end QA experiments using the passages retrieved by DPR.

\vspace{0.1cm}
\noindent\textbf{Ranking evaluation} measures the model's ability to differentiate a positive passage from negative passages, by ranking a set of candidate passages based on the relevance score to the question. We collect 50 candidates for each question, including a positive passage, 30 hard negative passages and 19 random negative passages. Hard negatives are the top-ranked passages in BM25 retrieval that do not contain the answer. We report Mean Rank (MR) and Mean Reciprocal Rank (MRR) of the positive passage.

\vspace{0.1cm}
\noindent\textbf{Retrieval evaluation} tests the model's ability to retrieve passages relevant to answering the question from a large corpus. Our retrieval corpus contains \textasciitilde21M passages from Wikipedia. We calculate Recall@$k$, the number of passages containing the answer in top-$k$ retrieved passages.

\vspace{0.1cm}
\noindent\textbf{End-to-end QA evaluation} checks whether the retrieved passages contain useful information for predicting the correct answer. The retrieved passages are fed into a Fusion-in-Decoder (FiD) reader~\cite{FiD} trained on NQ. We calculate Exact Match between model predictions and answers.

\section{Data: MEQ Contrast Sets}

\subsection{Dataset Construction}

Based on the above evaluation metrics, we collect two MEQ contrast sets to evaluate models' contrast consistency. The first set, referred to as MEQ-GPT, is generated using InstructGPT~\cite{instructgpt} then manually filtered and annotated with answers by crowdsource workers. The second set, named MEQ-AmbigQA, is sourced from the AmbigQA dataset~\cite{AmbigQA}. The construction of our contrast sets consists of four phases: question collection, MEQ filtering, answer annotation, and evidence passage annotation.

% The construction of our datasets consists of four phases. Firstly, we gather candidate MEQs from the aforementioned two sources. Secondly, we filter out questions that do not meet the criteria of MEQs. Thirdly, human annotators are employed to validate the quality of the remaining questions, while providing answers for GPT-generated questions. Finally, we again let crowdsource workers annotate gold evidence passages from Wikipedia for the collected questions.

\subsubsection{Collection of Candidate MEQs}

\paragraph{MEQ-InstructGPT}~{
Generating answerable MEQs is very difficult for crowdsource workers who are not domain experts. It is hard for them to determine which modifications to the original question result in an answerable MEQ without extensive Internet searches. However, recent GPT-3 models have demonstrated their ability to possess vast amount of knowledge through massive pre-training~\cite{GPT3}. Therefore, we first utilize the InstructGPT model (\textit{text-davinci-002}) to generate a set of MEQ candidates, and leave the answer annotation task to crowdsource workers. The input to InstructGPT is of the form $[I, x_1,\cdots, x_t, q, a]$, where $I$ is the instruction ``\texttt{Generate a similar question that has a different answer}''. $\{x_i\}^t_{i=1}$ are in-context demonstrations that are manually created, where each $x_i$ is a tuple $[q_i, a_i, q'_i, a'_i]$ ($q'_i$ is the MEQ of $q_i$). The original question $q$ and answer $a$ are appended to the input, prompting InstructGPT to generate a new question $q'$ and its answer $a'$ to complete the sequence. For each input $q$, we sample 10 completions from InstructGPT to generate a set of candidate MEQs.
}

\paragraph{MEQ-AmbigQA}~{
The AmbigQA dataset initially targeted a subset of NQ consisting of ambiguous questions. The dataset was introduced to decompose each ambiguous question into multiple disambiguated questions, each of which is a slight modification of the original question. For each NQ question covered in AmbigQA, its corresponding disambiguated questions are considered as its candidate MEQs and are delivered to the subsequent filtering phase (\S\ref{sec:MEQ_filter}).
However, such questions are limited as we set strict criteria for MEQs, so we need more data generated by InstructGPT for solid evaluation.
}

\subsubsection{MEQ Filtering}
\label{sec:MEQ_filter}
To build challenging contrast sets, a series of criteria are applied to eliminate unqualified candidates and select MEQs based on the definition in \S\ref{sec:formulation}. 
\begin{enumerate}
    \item \textbf{Quality control}: We do not allow $q$ and $q'$ to differ in question words (\textit{e.g.}, \texttt{how}, \texttt{what}), or if the only word that $q'$ adds to $q$ falls into \{\texttt{first,last,new,next,original,not}\}. We have found that InstructGPT frequently adds these words to create MEQs, but they usually lead to unanswerable questions.
    \item \textbf{Lexical distance}: Word-level edit distance is used as $d_{\ell}(q, q')$, and we remove $q'$ if $d_{\ell}(q, q') = 0\ \textrm{or}\ d_{\ell}(q, q') > 3$.
    \item \textbf{Semantic distance}: The cosine similarity of semantic embeddings is used to measure $d_s(q, q')$. We remove $q'$ if $\textrm{cos}(\mathbf{h}_q, \mathbf{h}_{q'}) < 0.95$ which indicates non-negligible semantic discrepancy. The semantic embedding $\mathbf{h}$ should be generated by a sentence embedding model. Here we use the question encoder of the unsupervised dense retrieval model Contriever~\cite{contriever}.
    \item \textbf{Paraphrase filtering}: $q'$ is discarded if it is determined as a paraphrase to $q$ by a paraphrase detection model. Here we use a RoBERTa-large~\cite{roberta} fine-tuned on the Quora Question Pairs dataset~\cite{GLUE} for paraphrase classification.
    \item \textbf{Answer Difference}: $q'$ is discarded if $a'=a$. For AmbigQA questions, since they are originally human-annotated, we ask human volunteers to check whether $a'$ and $a$ are aliases to the same entity. For GPT-generated questions, the inspection of answer difference is included in the answer annotation process, which we will elaborate in \S\ref{sec:annotate}.
 \end{enumerate}

 Among GPT-generated questions, for a certain original question $q$, there may be multiple MEQ candidates that pass the above filtering. In such cases, the question that is generated most frequently across 10 samples is selected as the most confident MEQ by InstructGPT. This is similar to the self-consistency idea in~\citet{self-consistency}.

% \usepackage{booktabs}
\begin{table}[t]
\resizebox{0.48\textwidth}{!}{
\setlength{\tabcolsep}{1.3mm}{
\begin{tabular}{l|cc|cc}
\toprule

&         \multicolumn{2}{c|}{\textbf{NQ}}                       & \multicolumn{2}{c}{\textbf{MEQ}} \\

\multirow{-2}{*}{\textbf{Statistics}}             & Train         & Test & AmbigQA & GPT \\ \midrule
Size                    & 79,168         & 3,610   & 1,050       & 2,293   \\
With Gold Passage   & 58,880      & 1,766       & 625         & 1,229   \\
Question Length  & 9.17        & 9.22    & 10.73       & 9.69    \\
Answer Length    & 2.16        & 2.22    & 2.62        & 1.96    \\
\#Answers         & 1.22       & 1.79    & 1.47        & 1.18    \\
Edit Distance   & 9.10      & 9.16       & \textbf{2.39}        & \textbf{1.18}    \\
Semantic Similarity & 30.12  & 29.87     & \textbf{96.47}       & \textbf{97.96}   \\ \bottomrule
\end{tabular}}}
\caption{Dataset statistics. Question lengths, answer lengths, and edit distances are all measured in \textit{words}. Semantic similarity is computed by Contriever~\cite{contriever}. For NQ-train and NQ-test, edit distance and semantic similarity are computed between random question pairs. For MEQ contrast sets, they are computed between the original question and its MEQ.}
\label{tab:data_stats}
\end{table}
\subsubsection{Answer Annotation}
\label{sec:annotate}

Due to the limited accuracy of InstructGPT in directly answering open-domain questions~\cite{Genread}, we recruit crowdsource workers to annotate the answer of each candidate MEQ generated by InstructGPT. Before human annotation, we first check the answer generated by InstructGPT via Google Search. If Google Search returns an highlighted answer box which matches the InstructGPT-generated answer, we skip the subsequent human labeling step.  For the remaining questions, we recruit human annotators from Surge AI\footnote{\url{https://www.surgehq.ai}} for data labeling. We ask them the following questions:
\begin{enumerate}
    \item[\textbf{Q1.}] Is $q'$ a good variation to $q$? Bad variations include being unanswerable or having the same answer with $q$, and are discarded from our dataset.
    \item[\textbf{Q2.}] If $q'$ is deemed a good variation, find the answer $a'$ using search engines. If necessary, the question may have multiple answers.
\end{enumerate}

% It is worth noting that \textbf{Q1} includes filtering out candidate MEQs where $a'=a$. For MEQ-AmbigQA, as AmbigQA is originally human-annotated, we only need to check whether $a'$ and $a$ are aliases to the same entity, which is done by human volunteers.

\paragraph{Quality control}~{To ensure answer correctness, each question is answered by two different annotators. If the annotators disagree on the answer or if either annotator determines the question is an bad variation, the question is discarded. Since the answers are free-form responses, we manually check whether the answers given by two annotators are aliases to the same entity. If the response of the first annotator matches exactly with the answer provided by InstructGPT, we do not recruit a second annotator to reduce costs.}

% Please add the following requicrimson packages to your document preamble:
% \usepackage{booktabs}
\begin{table*}[t]
\centering
\setlength{\tabcolsep}{1.5mm}
\resizebox{0.95\textwidth}{!}{
\begin{tabular}{cccl}
\toprule
\textbf{Edit Type}                                                            & \textbf{Proportion} & \textbf{Antonym Edits} & \multicolumn{1}{c}{\textbf{Example}}                                                                                                                                                                                                                       \\ \midrule
Nouns                                                                & 641 (28.0\%)  & 151 (24\%)     & \begin{tabular}[c]{@{}l@{}}   Q: Who wrote the {\color{ForestGreen}music} for the national anthem? A:  John Stafford Smith\\  Q: Who wrote the {\color{crimson}lyrics} for the national anthem? A:  Francis Scott Key\end{tabular}              \\\midrule
Verbs                                                                & 425 (18.5\%)  & 176 (41\%)     & \begin{tabular}[c]{@{}l@{}} Q: When did Australia {\color{ForestGreen}stop} using one cent coins? A:  1992\\ Q: When did Australia {\color{crimson}start} using one cent coins? A:  1966\end{tabular}                                             \\\midrule
Adjectives                                                           & 418 (18.2\%)  & 146 (35\%)     & \begin{tabular}[c]{@{}l@{}} Q: How many islands are in Andaman and Nicobar? A:  572\\ Q: How many {\color{crimson}inhabited} islands are in Andaman and Nicobar? A:  37\end{tabular}                                              \\\midrule
Numbers                                                              & 326 (14.2\%)  & -     & \begin{tabular}[c]{@{}l@{}}Q: Where did season {\color{ForestGreen}2} of Jersey Shore take place?  A:  Miami Beach, Florida\\  Q: Where did season {\color{crimson}3} of Jersey Shore take place? A:  Seaside Heights, New Jersey\end{tabular}  \\\midrule

Ordinals                                                             & 211 (9.2\%)  & 30  (14\%)     & \begin{tabular}[c]{@{}l@{}} Q: Highest scoring NBA players of all time in one game? A:  Wilt Chamberlain\\  Q: {\color{crimson}Second} highest scoring NBA players of all time in one game? A:  Kobe Bryant\end{tabular} \\\midrule

Dates                                                                & 152 (6.6\%)  & -     & \begin{tabular}[c]{@{}l@{}} Q: Who ruled the Holy Roman Empire in {\color{ForestGreen}1509}? A:  Maximilian I\\  Q: Who ruled the Holy Roman Empire in {\color{crimson}1519}? A:  Charles V\end{tabular}                                      \\\midrule

\begin{tabular}[c]{@{}c@{}}Prepositions\\ Conjunctions\end{tabular} & 66 (2.9\%)  & 14 (21\%)     & \begin{tabular}[c]{@{}l@{}} Q: Where did the Titanic make its maiden voyage {\color{ForestGreen}from}? A:  Southampton\\  Q: Where did the Titanic make its maiden voyage {\color{crimson}to}? A:  New York\end{tabular}                      \\ \bottomrule
\end{tabular}}
\caption{Different MEQ edit types in MEQ-GPT with their proportions of antonym edits and examples. The remaining 2.4\% of the instances are of miscellaneous types. The first line in each example is the original question and the second line is the MEQ. Words in {\color{ForestGreen}green} and {\color{crimson}red} are the deleted and added words, respectively.}
\label{tab:edit_types}
\end{table*}

\subsubsection{Gold Evidence Passages}

As mentioned in \S\ref{sec: evaluation}, ranking evaluation on MEQs needs gold evidence passages as positive examples, so we collect them from Wikipedia for our contrast sets. For MEQ-AmbigQA, we utilize the semi-oracle evidence documents\footnote{\url{https://github.com/shmsw25/AmbigQA/blob/main/evidence.md}} provided by the original authors, dividing them into 100-word passages. Then, we identify the first passage that contains the gold answer. For MEQ-GPT, our initial step involves finding candidate evidence passages that include the gold answer. This is achieved by retrieving Wiki passages with BM25 and selecting the top 3 passages that contain the answer. Next, we recruit human annotators from Surge AI to assess whether any of these passages provide sufficient evidence for answering the question. The highest-ranked passage that passed human annotation is chosen as the gold evidence passage. Finally, both contrast sets have a subset of questions paired with a corresponding gold evidence passage.

\subsection{Dataset Analysis}

The full dataset is composed of 3,343 MEQs (2,293 from InstructGPT and 1,050 from AmbigQA). Each of these MEQs has its original question in the NQ training set. Among them, 1,229 (53.6\%) InstructGPT questions and 625 (59.5\%) AmbigQA questions are paired with a gold evidence passage from Wikipedia. We use this subset in ranking evaluation and the full set in retrieval and end-to-end QA evaluation.

\paragraph{Data statistics}~{
We summarize basic statistics of the MEQ contrast sets compared to the original NQ questions. As shown in Table~\ref{tab:data_stats}, MEQ-GPT is similar to NQ regarding the average length of questions and answers. Questions in MEQ-AmbigQA are longer because the original AmbigQA annotators usually added conditions to disambiguate the original NQ questions. Besides, AmbigQA does not impose a limit on the answer length, while we limit each answer in MEQ-GPT to at most 5 words, consistent with NQ. The number of answers per question is lower in MEQ-GPT than in MEQ-AmbigQA, because most answers are obtained through strict text matching on candidate answers from two sources. In addition, we observe that MEQ-GPT has a smaller edit distance and higher semantic similarity between $q$ and $q'$, making it hard for models to distinguish them.
}

\begin{figure}
    \centering
    \includegraphics[width=0.4\textwidth]{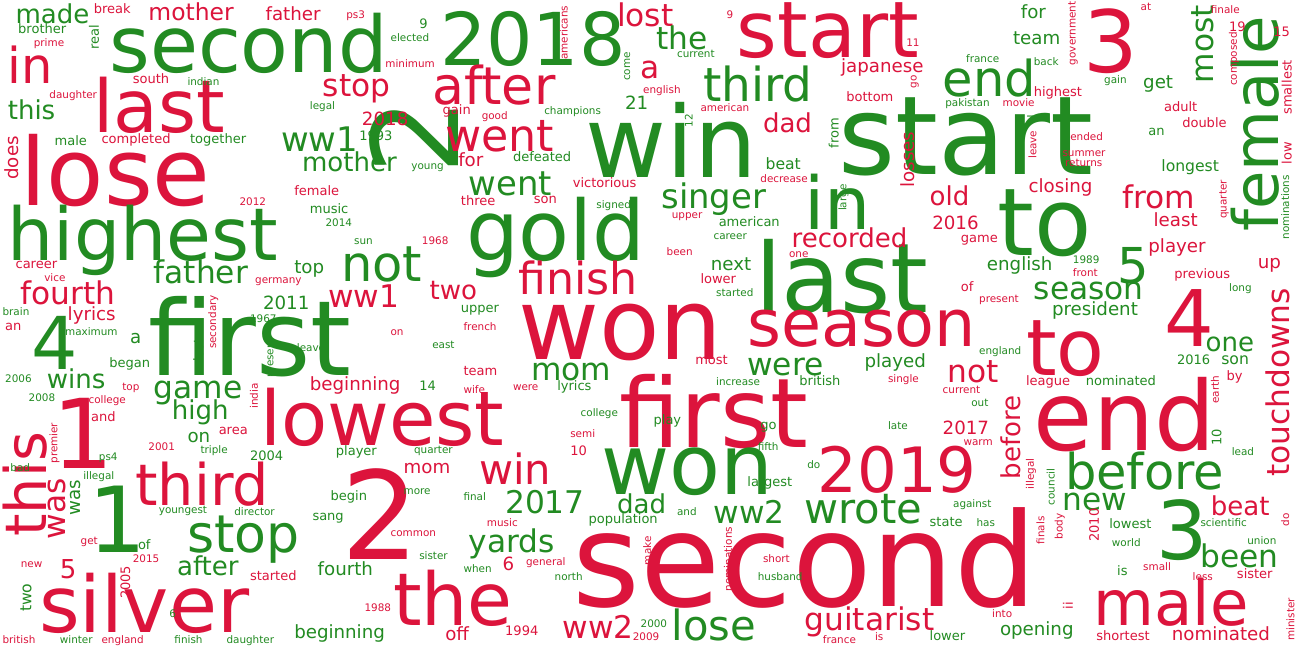}
    \caption{Word cloud of the edited words. Words in {\color{ForestGreen}green} and {\color{crimson}red} are the deleted and added words, respectively. Larger font sizes indicate higher frequencies.}
    \label{fig:word_cloud}
\end{figure}

\paragraph{Types of edits}~{
We review and categorize different types of minimal edits that are used to create MEQs. Since MEQ-AmbigQA primarily consists of edits that add specifications to the original NQ question, we consider MEQ-GPT as a more natural representation of minimal edits. As shown in Table~\ref{tab:edit_types}, the edits in MEQ-GPT involve nouns (28.0\%), verbs (18.5\%), adjectives (18.2\%), numbers (14.2\%), ordinals (9.2\%), dates (6.6\%), prepositions/conjunctions (2.9\%) and others (2.4\%). A word cloud of the edited words is given in Figure~\ref{fig:word_cloud}. We also observe that 22.5\% of the total edits are antonym edits where a word in the original question is replaced by its antonym. Our dataset of diverse MEQs provides a comprehensive evaluation of contrast consistency.
}

\subsection{Challenges of MEQ Contrast Sets}
\label{sec:MEQ_difficulty}

The collected MEQ contrast sets are challenging for the widely-used DPR-based OpenQA system, although these perturbed questions are only minimal edits to the well-learned training questions. As shown in Figure~\ref{fig:example}, the model significantly underperforms on the contrast sets, where the passage ranking score of DPR decreases by 39\% and 45\% compared to NQ-train, and by 29\% and 18\% compared to NQ-test. This makes a substantial impact on the QA performance, with the accuracy being 69\% and 60\% lower on the two contrast sets compared to NQ-train, and 54\% and 40\% lower than NQ-test. The results show that the collected MEQs are much harder to solve than random test questions, which indicates our contrast sets can serve as testbeds for evaluating the contrast consistency of OpenQA.

% \section{Method: Improving DPR's Prediction Consistency}

\section{Method: Training DPR with Query- Side Contrastive Loss}

\begin{figure*}[t]
    \centering
    \includegraphics[width=0.9\textwidth]{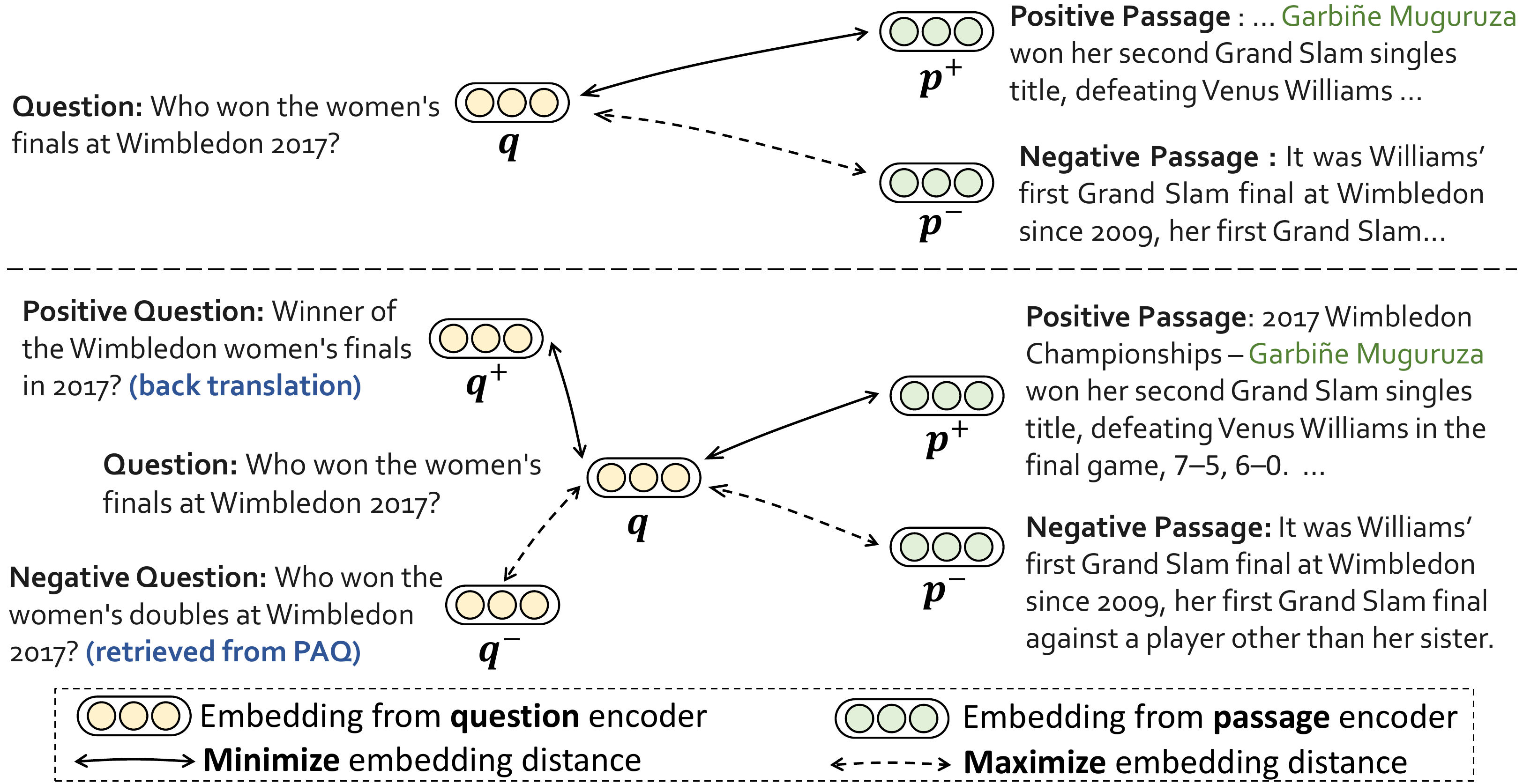}
    \caption{Above: the original contrastive training of DPR. Below: our improved DPR with the query-side contrastive loss, where $q^{+}$ and $q^{-}$ are obtained through data augmentation.}
    \label{fig:framework}
\end{figure*}

\subsection{Preliminary: DPR}

As a dense retriever, DPR includes a question encoder $E_{Q}(\cdot)$ and a passage encoder $E_{P}(\cdot)$. Both encoders map the input sequence to a dense vector as its semantic representation. The relevance score $s(q,p)$ between a question $q$ and a passage $p$ is defined as the dot product of their representations:
\begin{equation}
\nonumber
    s(q,p) = {E_{Q}(q)}^{\intercal}E_{P}(p)
\end{equation}

DPR is trained via a contrastive loss. Given a positive passage $p^{+}$ and a set of negative passages $\{p^{-}_i\}^{n}_{i=1}$ to a certain question, the model is trained to maximize the relevance score between $q$ and $p^{+}$, while minimizing the relevance score between $q$ and each $p^{-}_i$.
% Negative passages usually include random in-batch negatives, as well as hard negatives which are top-ranked passages retrieved by BM25 without containing the answer.
The loss function is:
\begin{equation}
\nonumber
    % &L(q,p^{+},\{p^{-}\}^{n}_{i=1}) =\\ -&\textrm{log}\frac{\textrm{exp}(s(q,p^{+})/\tau)}{\textrm{exp}(s(q,p^{+})/\tau) + \sum^{n}_{i=1}\text{exp}(s(q,p^{-}_i)/\tau)}
    % &L(q,p^{+},\{p^{-}\}^{n}_{i=1}) =\\ -&\textrm{log}\frac{\textrm{exp}(s(q,p^{+}))}{\textrm{exp}(s(q,p^{+})) + \sum^{n}_{i=1}\text{exp}(s(q,p^{-}_i))}
    \mathcal{L}_{QP} = -\textrm{log}\frac{\textrm{exp}(s(q,p^{+}))}{\textrm{exp}(s(q,p^{+})) + \sum^{n}_{i=1}\text{exp}(s(q,p^{-}_i))}
\end{equation}

The above training paradigm works well on retrieving passages for random test questions, but does not perform as effectively on MEQ contrast sets, as discussed in \S\ref{sec:introduction} and  \S\ref{sec:MEQ_difficulty}.
% In an \textit{i.i.d.} training set, questions are distant from each other in the data space. Moreover, 
The training loss $\mathcal{L}_{QP}$ does not provide explicit signals for DPR to learn the relationships between questions. As a result, the question embeddings are insensitive to minimal discrepancies, which prevents the model from identifying the MEQ as a distinct question after seeing the original question in training. This causes DPR to generate an overly similar embedding for the MEQ, leading to a high overlap in the retrieved passages and low contrast consistency.

% During inference, all Wikipedia passages are encoded as embeddings. The FAISS~\cite{faiss} library is used to build an index on all passages and conduct a nearest neighbor search to find the most relevant passages to the question. We refer readers to~\cite{DPR} for further details of DPR.

\subsection{Proposed Method}

We propose to improve the contrast consistency of DPR by introducing a query-side contrastive loss to distinguish between paraphrase questions and MEQs which are positive and negative question examples for an original question, respectively.
We devise a data augmentation approach to collect synthetic question examples to train this loss.
% The augmented MEQs are also subject to learning question-passage relationships

\subsubsection{Data Augmentation}
\label{sec:data_augment}

For a training question $q$, its positive example $q^{+}$ is a synthetic paraphrase question which is slightly different from $q$ and has the same answer; its negative question $q^{-}$ is a synthetic MEQ with a different answer.

To obtain $q^{+}$, we leverage back translation provided by the \texttt{nlpaug}\footnote{\texttt{https://github.com/makcedward/nlpaug}} package. The original question $q$ is translated to another language and then translated back to produce a new phrasing of $q$. We used translation models of 6 languages provided by~\citet{WMT19} and~\citet{OPUS}. Questions that are identical to $q$ (\textit{i.e.}, edit distance = 0) or classified as ``not paraphrase'' by the paraphrase detection model used in \S\ref{sec:MEQ_filter} are eliminated. The remaining questions constitute a candidate set of positive questions from which a random $q^{+}$ is sampled in each epoch.

To obtain $q^{-}$, synthetic MEQs are retrieved from the machine-built QA corpus PAQ~\cite{PAQ}. All questions in PAQ that are similar to $q$ are retrieved by the question retriever in the work of PAQ. Then, the MEQ requirements specified in \S\ref{sec:MEQ_filter} are applied to filter the retrieved synthetic questions. The remaining questions constitute a candidate set of negative questions from which a random $q^{-}$ is sampled in each epoch.

Apart from learning the relationships among $q$, $q^{+}$, and $q^{-}$, the loss $\mathcal{L}_{QP}$ can be augmented to learn the relevance between synthetic questions and their corresponding passages. Because $q^{+}$ is a paraphrase question mapping the passages of $q$, it does not have to be involved in $\mathcal{L}_{QP}$. To train on $q^{-}$, its positive passage is the Wikipedia passage that was used to generate the question during the construction of PAQ; its negative passages are collected from the top-ranked passages retrieved by BM25 which do not contain the answer. 
% Besides $\mathcal{L}_{QP}$, $q^{-}$ is also jointly trained with $\mathcal{L}_{QQ}$, just as the original training questions. The positive question of $q^{-}$ is colleted through back translation, and negative question of $q^{-}$ is exactly $q$ due to the symmetry of the MEQ relationship.

% \usepackage{multirow}
\begin{table*}[t]
\centering
\setlength{\tabcolsep}{3.5mm}{
\resizebox{0.875\textwidth}{!}{
\begin{tabular}{c|c|cc|cc|cc}
\toprule
                        &                                                                 & \multicolumn{2}{c|}{\textbf{NQ}}                                                       & \multicolumn{2}{c|}{\textbf{MEQ-AmbigQA}}                                              & \multicolumn{2}{c}{\textbf{MEQ-GPT}}                                                \\
\multirow{-2}{*}{\textbf{Model}} & \multirow{-2}{*}{\textbf{Augmentation}}                 & \multicolumn{1}{c}{\textbf{MR$\downarrow$}}               & \multicolumn{1}{c|}{\textbf{MRR$\uparrow$}}               & \multicolumn{1}{c}{\textbf{MR$\downarrow$}}               & \multicolumn{1}{c|}{\textbf{MRR$\uparrow$}}               & \multicolumn{1}{c}{\textbf{MR$\downarrow$}}               & \multicolumn{1}{c}{\textbf{MRR$\uparrow$}}               \\
\midrule
$\text{DPR}_{\text{BASE}}$              & None                           & 2.36                                 & 0.784                                 & 5.09                                 & 0.563                                 & 5.44                                 & 0.507                                 \\
$\text{DPR}_{\text{BASE}}$              & Random                             & 2.36                                 & 0.781                                 & 5.09                                 & 0.557                                 & 5.25                                 & 0.524                                 \\
$\text{DPR}_{\text{BASE}}$              & MEQs                              & 2.34                        & 0.783                        & 5.09                                 & 0.543                                 & 5.10                                 & 0.529                                 \\
$\text{DPR}_{\text{BASE}}$              & MEQs              + $\mathcal{L}_{QQ}$ & \textbf{2.25}                                 & \textbf{0.791}                                 &  \textbf{4.85} &  \textbf{0.569} & \textbf{4.88}                        & { \textbf{0.547}} \\ \midrule
$\text{DPR}_{\text{LARGE}}$             & None                                          & 2.31                                 & 0.780                                 & 4.84                                 & 0.569                                 & 5.46                                 & 0.515                                 \\
$\text{DPR}_{\text{LARGE}}$             & Random                                      & 2.20                                 & 0.797                                 & 4.98                                 & 0.554                                 & 5.18                                 & 0.533                                 \\
$\text{DPR}_{\text{LARGE}}$             & MEQs                                   & 2.17                                 & 0.797                                 & 4.79                                 & 0.561                                 & 5.00                                 & 0.544                                 \\
$\text{DPR}_{\text{LARGE}}$             & MEQs                  + $\mathcal{L}_{QQ}$ &  \textbf{2.14} &  \textbf{0.804} &  \textbf{4.59} &  \textbf{0.592} &  \textbf{4.61} &  \textbf{0.565} \\
\bottomrule
\end{tabular}}}
\caption{Ranking evaluation results. MR and MRR stand for mean rank and mean reciprocal rank, respectively. A lower MR or higher MRR indicates better performance. BM25 is not listed because sampling hard negatives from top-ranked passages in BM25 retrieval lowers the ranking performance of BM25 in return.}
\label{tab:ranking}
\end{table*}
% \usepackage{multirow}
\begin{table}[t]
\resizebox{0.48\textwidth}{!}{
\setlength{\tabcolsep}{1.5mm}{
\begin{tabular}{c|c|cc|cc}
\toprule                      &                                                                                            & \multicolumn{2}{c|}{\textbf{AmbigQA}}                                              & \multicolumn{2}{c}{\textbf{GPT}}                                                \\
\multirow{-2}{*}{\textbf{Model}} & \multirow{-2}{*}{\textbf{$\mathcal{L}_{QQ}$}}                            & \multicolumn{1}{c}{\textbf{MR$\downarrow$}}               & \multicolumn{1}{c|}{\textbf{MRR$\uparrow$}}               & \multicolumn{1}{c}{\textbf{MR$\downarrow$}}               & \multicolumn{1}{c}{\textbf{MRR$\uparrow$}}               \\
\midrule
$\text{DPR}_{\text{BASE}}$              & InfoNCE                                                       & 4.85                                 & 0.569                                 & \textbf{4.88}                                 & \textbf{0.547}                                 \\
$\text{DPR}_{\text{BASE}}$              & Dot Product                                                           & \textbf{4.79}                                 & \textbf{0.574}                                 & 4.98                                 & 0.539                                 \\
$\text{DPR}_{\text{BASE}}$              & Triplet                         & 4.80                        &  0.568                        & 4.91                                 & 0.542                                                                 \\ \midrule
$\text{DPR}_{\text{LARGE}}$             & InfoNCE                                                                        & 4.76                                 & 0.572                                 & 4.63                                 & 0.570                                 \\
$\text{DPR}_{\text{LARGE}}$             & Dot Product                                                         & \textbf{4.59}                                 & \textbf{0.592}                                 & 4.61                                 & 0.565                                 \\
$\text{DPR}_{\text{LARGE}}$             & Triplet                                                                & 4.61                                 & 0.582                                 & \textbf{4.59}                                 & \textbf{0.573}                              \\
\bottomrule
\end{tabular}}}
\caption{Ranking evaluation with different $\mathcal{L}_{QQ}$ functions on two MEQ contrast sets. All loss functions outperform the baselines in Table~\ref{tab:ranking}.}
\label{tab:contrast_loss}
\end{table}

\subsubsection{Model Training}
\label{sec:query-loss}

To provide more supervision signals and prevent overfitting, we randomly sample $q^{+}$, $q^{-}$, and $p^{-}$ for each training question $q$ in each epoch. This means while the original training questions remain fixed, a different set of augmented questions is used. For explicit supervision on inter-question relationships, given $q$, DPR is trained to assign higher relevance scores to its paraphrase question ($q^{+}$) and lower relevance scores to its MEQ ($q^{-}$). The relevance score of any pair of questions $(q_1, q_2)$ is calculated as the inner product of their embeddings:
$s(q_1,q_2) = {E_{Q}(q_1)}^{\intercal}E_{Q}(q_2)$.
Specifically, we consider three forms of query-side constrastive loss functions in experiments:

\vspace{0.1cm}
\textbf{(1) InfoNCE Loss}~\cite{InfoNCE}, which differentiates the positive question from a set of $m$ negative questions. Besides the synthetic MEQ which is considered as a hard negative, the other questions in the same batch are included as random negatives. The loss function is:
\begin{equation}
\nonumber
    \mathcal{L}_{QQ} = -\textrm{log}\frac{\textrm{exp}\left(s\left(q,q^{+}\right)\right)}{\textrm{exp}(s(q,q^{+})) + \sum^{m}_{j=1}\text{exp}(s(q,q^{-}_j))}.
\end{equation}

\textbf{(2) Dot Product Loss}, which directly penalizes the relevance score between a sample question $q$ and its augmented MEQ:
\begin{equation}
\nonumber
    \mathcal{L}_{QQ} = s(q,q^{-}).
\end{equation}

\textbf{(3) Triplet Loss}~\cite{Triplet}, which trains the model to assign a higher relevance score to $q^{+}$ compared to $q^{-}$, enfored by a margin $\alpha$:
\begin{equation}
\nonumber
    \mathcal{L}_{QQ} = \textrm{max}\left(0, \alpha - s(q, q^{+}) + s(q, q^{-})\right).
\end{equation}

The final training loss of our improved DPR is $\mathcal{L}=\mathcal{L}_{QP}+\lambda\mathcal{L}_{QQ}$, where the hyperparameter $\lambda$ weights the trade-off between the loss terms.

% (1) q-也找到自己的q+, q-, p+, p-，然后像传统data augmentation那样放进训练集中，其中q-的q-就是q (q+就不用了因为q+和q是一个question）
% (2) 为了有更多的signal，q+, q-, p-都是每个epoch随机从candidates中sample一个

\section{Experiments}

In experiments, we compare our proposed training method against the original training setting of DPR. After training the models on the NQ training set, we test them on the standard NQ test set as well as two MEQ contrast sets that we collected in this work. 

% \usepackage{multirow}
\begin{table*}[t]
\centering
\setlength{\tabcolsep}{2.5mm}{
\resizebox{0.9\textwidth}{!}{
\begin{tabular}{c|c|ccc|ccc|ccc}
\toprule
                        &                                & \multicolumn{3}{c|}{\textbf{NQ}}                                                                                                    & \multicolumn{3}{c|}{\textbf{MEQ-AmbigQA}}                                                                                                        &                                                           \multicolumn{3}{c}{\textbf{MEQ-GPT}}                                                                                                                                                                     \\
\multirow{-2}{*}{\textbf{Model}} & \multirow{-2}{*}{\textbf{Augmentation}} & \textbf{R@1}                              & \textbf{R@5}                              & \textbf{R@20}                             & \textbf{R@1}                                  & \textbf{R@5}                                  & \textbf{R@20}                                                                                        & \textbf{R@1}                                  & \textbf{R@5}                                  & \textbf{R@20}       
\\ \midrule
BM25              & None                           & 23.2                             & 45.3                             & 64.5 & 16.8                                 & 34.2                                 & 48.8                                                           & 21.1                                 & 42.7                                 & 61.7 \\ \midrule
$\text{DPR}_{\text{BASE}}$              & None                           & 46.6                             & 70.0                             & 81.2 & 28.5                                 & 50.0                                 & 65.6                                                           & 31.5                                 & 57.3                                 & 73.2                                          \\
$\text{DPR}_{\text{BASE}}$              & Random                         & \textbf{48.2}                             & \textbf{71.2}                             & 81.6  & 27.5                                 & 49.2                                 & 65.8                                                             & 31.8                                 & 58.0                                 & 73.8                                                         \\
$\text{DPR}_{\text{BASE}}$              & MEQs                           & 46.4                        & 69.9                        &     81.3                                    & 25.2                                 & 46.2                                 & 62.7                                                 & 31.5                                 & 55.9                                 & 72.3                                              \\
$\text{DPR}_{\text{BASE}}$              & MEQs + $\mathcal{L}_{QQ}$     &       48.1                           &            70.8                      &       \textbf{81.9}                      & \textbf{29.5} & \textbf{52.3} & \textbf{66.4}& \textbf{32.8} & \textbf{58.7} & \textbf{74.4}  \\ \midrule
$\text{DPR}_{\text{LARGE}}$             & None                           &                46.0                  &        67.6                                &      80.3       &      26.8                                &    49.2                                  &                  64.0                                       &       29.2                               &                  54.9                    &           70.7                                                                                                                                              \\
$\text{DPR}_{\text{LARGE}}$             & Random                         &           49.0                       &      70.9                            &          81.5            &        26.2                              &                            48.4          &             64.1                                                                                          &        31.3                              &                    56.7                  &               72.3                                                                                    \\
$\text{DPR}_{\text{LARGE}}$             & MEQs                           &     48.0          &              70.5                        &                       81.4                                                     &      27.7                  &                 47.2                 &          61.5                      &     31.2                                 &                       57.0               &      71.8                                                                                                                                                          \\
$\text{DPR}_{\text{LARGE}}$             & MEQs + $\mathcal{L}_{QQ}$     &  \textbf{51.0}  &       \textbf{71.2}                   &             \textbf{81.6}             &   \textbf{30.1}    &   \textbf{52.3}    &   \textbf{65.4}                                     &       \textbf{32.5}                                                   &  \textbf{58.4}     &  \textbf{73.1}                                              \\ \bottomrule                                                    
\end{tabular}}}
\caption{Retrieval evaluation results. R@$k$ stands for Recall@$k$.}
\label{tab:retrieval}
\end{table*}

\begin{table*}[t]
\centering
\resizebox{0.9\textwidth}{!}{
\setlength{\tabcolsep}{3mm}{

\begin{tabular}{c|c|ccc|ccc|ccc}
\toprule
                        &                                & \multicolumn{3}{c|}{\textbf{NQ}}                                                                                                    & \multicolumn{3}{c|}{\textbf{MEQ-AmbigQA}}                                                                                                        &                                                           \multicolumn{3}{c}{\textbf{MEQ-GPT}}                                                                                                                                                                     \\
\multirow{-2}{*}{\textbf{Model}} & \multirow{-2}{*}{\textbf{Augmentation}} & 1P                              & 5P                              & 20P                  & 1P                              & 5P                              & 20P            & 1P                              & 5P                              & 20P  
\\ \midrule
BM25              & None                           &     16.4                         &            28.4                  & 37.3 &       10.9                           & 15.2                                 & 18.1                                                              & 13.3                                 & 20.5   & 25.8       \\ \midrule
$\text{DPR}_{\text{BASE}}$              & None                           &     32.6                         &            43.2                  & 49.1 &       14.0                           & 19.7                                 & 21.9                                                              & 17.6                                 & 25.8                                 & 29.3                                                     \\
$\text{DPR}_{\text{BASE}}$              & Random                         &     33.7                         &            \textbf{44.8}                  & \textbf{49.4}  & 14.7                                 & 19.1                                 & 22.4                                                         & 16.8                                 & 25.4                                 & 29.5                                                   \\
$\text{DPR}_{\text{BASE}}$              & MEQs                           & 32.0                        & 43.4                        &    48.7                                     & 13.5                                 & 19.3                                 & \textbf{23.1}                                                            & 17.1                                 & 25.5                                 & 29.5                                                      \\
$\text{DPR}_{\text{BASE}}$              & MEQs + $\mathcal{L}_{QQ}$     &      \textbf{34.4}                            &          44.7                        &            49.2                              & \textbf{16.6} & \textbf{21.8} & 22.8& \textbf{19.5} & \textbf{26.7} & \textbf{31.1}\\ \midrule
$\text{DPR}_{\text{LARGE}}$             & None                           &              31.4                              &     42.2                     &         47.9                 &            14.3                          &        19.2                              &                 21.4                                                                                      &                   16.1                   &     24.6                                 &                29.1                                                                     \\
$\text{DPR}_{\text{LARGE}}$             & Random                         &               \textbf{33.7}                   &         44.6                                 &        \textbf{49.3}                  &             13.4                         &       20.4                               &             21.5                                                                                   &                17.3                      &    25.5                                  &              29.4                                                                         \\
$\text{DPR}_{\text{LARGE}}$             & MEQs                           &              33.0                    &          \textbf{44.7}                           &          48.7                &        15.7                              &          19.3                            &             21.7                                                                                         &          17.4                            &   25.0                                   &              29.1                                                                         \\
$\text{DPR}_{\text{LARGE}}$             & MEQs + $\mathcal{L}_{QQ}$     &   \textbf{33.7}  &        44.6                  &        \textbf{49.3}                  &  \textbf{16.1}     &  \textbf{22.1}     &                                 \textbf{23.0}                                                                    &  \textbf{19.4}     &   \textbf{27.6}    &                                        \textbf{31.6}                 \\ \bottomrule                                                    
\end{tabular}}}
\caption{End-to-end QA results (Exact Match). 1P, 5P and 20P are the number of passages read by the FiD reader.}
\label{tab:qa}
\end{table*}

\subsection{Models}
We augment the training set with $M$=33k synthetic MEQs and train DPR with both $\mathcal{L}_{QP}$ and $\mathcal{L}_{QQ}$. We consider the following baselines:
\begin{itemize}
    \item \textbf{Vanilla DPR}. This is the original training setting of DPR, proposed by~\cite{DPR}. The model is trained only with $\mathcal{L}_{QP}$ on the standard NQ training set.
    \item \textbf{DPR with random augmented questions}. This model is trained only with $\mathcal{L}_{QP}$, but we add $M$ random synthetic questions from PAQ to the training set. This is to rule out the effect of simply adding more synthetic data.
    \item \textbf{DPR with augmented MEQs}. This model uses the same set of $M$ synthetic MEQs retrieved from PAQ as data augmentation, but is trained only with $\mathcal{L}_{QP}$. We use this variant to test if $\mathcal{L}_{QQ}$ is necessary in model training.
\end{itemize}
Besides, we test the performance of BM25 on retrieval as a reference. Recent research has shown that larger retrievers may exhibit better generalization~\cite{Large_T5_retrievers}. Therefore, in addition to the standard DPR which is built on BERT-Base, we use BERT-Large as the backbone model to see: (1) whether MEQ contrast sets are still challenging for larger models and (2) whether our training method is still effective for larger models. We name the smaller model and larger model $\text{DPR}_{\text{BASE}}$ and $\text{DPR}_{\text{LARGE}}$, respectively.

We use the same set of basic hyper-parameters for each DPR model: a learning rate of $10^{-5}$, a batch size of 64 (32 for $\text{DPR}_{\text{LARGE}}$), 40 training epochs with 5\% warmup steps. On ranking evaluation, our best setting uses the InfoNCE loss with $\lambda=0.5$. On retrieval and QA evaluation, our best setting uses the dot product loss with $\lambda=0.03$. Since we do not have a dev set for MEQs,\footnote{We empirically found that model performance on NQ-dev is inconsistent with MEQ contrast sets.} we conduct ranking evaluation on MEQ contrast sets in a dev setting, where we select the highest score among all checkpoints. Then we use the checkpoint with the best ranking score to test its retrieval and QA performance. The scores on NQ-test is reported using the best checkpoint on NQ-dev.

\subsection{Results}

Experimental results on three datasets (NQ-test, MEQ-AmbigQA, MEQ-GPT) are presented from Table \ref{tab:ranking} to Table \ref{tab:qa}. We have the following findings:

\vspace{0.03in}
\noindent \textbf{(1) Our proposed method improves DPR's ability to distinguish MEQs.} As shown in Tables~\ref{tab:ranking} and~\ref{tab:retrieval}, on passage ranking and passage retrieval, the DPR trained with query-side contrastive loss outperforms the vanilla DPR on both contrast sets, showing improved contrast consistency on MEQs. This improvement is consistent across models of different sizes. For example, on MEQ-GPT, our model improves the vanilla DPR by 8\% and 10\% on ranking MRR for base and large versions respectively. On the choice of $\mathcal{L}_{QQ}$, Table~\ref{tab:contrast_loss} demonstrates that all three loss functions improve performance over baselines, while the optimal setting may require tuning on the specific dataset.

\vspace{0.1cm}
\noindent  \textbf{(2) The query-side contrastive loss contributes the most to the improved contrast consistency.} Although synthetic MEQs themselves bring more training signals, the model cannot consistently outperforms the vanilla DPR without $\mathcal{L}_{QQ}$. Actually, its performance is sometimes even lower than DPR. In contrast, after including the query-side contrastive loss, we observe consistent improvements across all datasets, as shown in Tables~\ref{tab:ranking} and~\ref{tab:retrieval}. For example, on MEQ-AmbigQA, simply adding synthetic MEQs into the training set gives 12\% lower recall@1 than the vanilla DPR, while training with $\mathcal{L}_{QQ}$ outperforms the naive augmentation method by 18\%.

\vspace{0.1cm}
\noindent \textbf{(3) The improvement does not simply come from the increased number of training data.} There is no significant difference on the performance between DPR augmented with random synthetic questions (``Random'' in ``Augmentaiton'' column) and the original DPR (``None'' in the column) in Tables~\ref{tab:ranking},~\ref{tab:retrieval}, and~\ref{tab:qa}. The average improvement of inserting random synthetic questions on all metrics is only 0.2\% for $\text{DPR}_{\text{BASE}}$ and 1.6\% for $\text{DPR}_{\text{LARGE}}$, which indicates simply adding more synthetic data is not an effective solution.

\vspace{0.1cm}
\noindent \textbf{(4) Improved retrieval performance leads to higher end-to-end QA accuracy.} As shown in Table~\ref{tab:qa}, our improved DPR provides more relevant information for answer prediction on MEQs. Even using only 1 retrieved passage, our improved DPR-Large outperforms its vanilla version by 12\% and 11\% on two contrast sets respectively.

\vspace{0.1cm}
\noindent \textbf{(5) Our method does not sacrifice performance on standard test questions.} After jointly trained with the query-side contrastive loss and augmented with synthetic MEQs, our model still maintains its competitive performance on the standard NQ test set. Specifically, It outperforms all baselines in ranking evaluation (see Table~\ref{tab:ranking}), while performing on par with the best baseline in retrieval and QA scores (see Tables~\ref{tab:retrieval} and \ref{tab:qa}).

\textbf{Summary}: The results are consistent across ranking, retrieval, and end-to-end QA experiments, which demonstrates the solidity of the above findings. Nevertheless, the performance of DPR still has a long way to improve, and such a gap is observed in both base and large versions of the model. Notably, DPR models perform significantly worse on MEQ contrast sets than the standard test set, even though it is trained under a development setting. This suggests that further research is still necessary to improve the contrast consistency of retrieval models on MEQs.

\begin{figure}[t]
    \centering
    \includegraphics[width=0.48\textwidth]{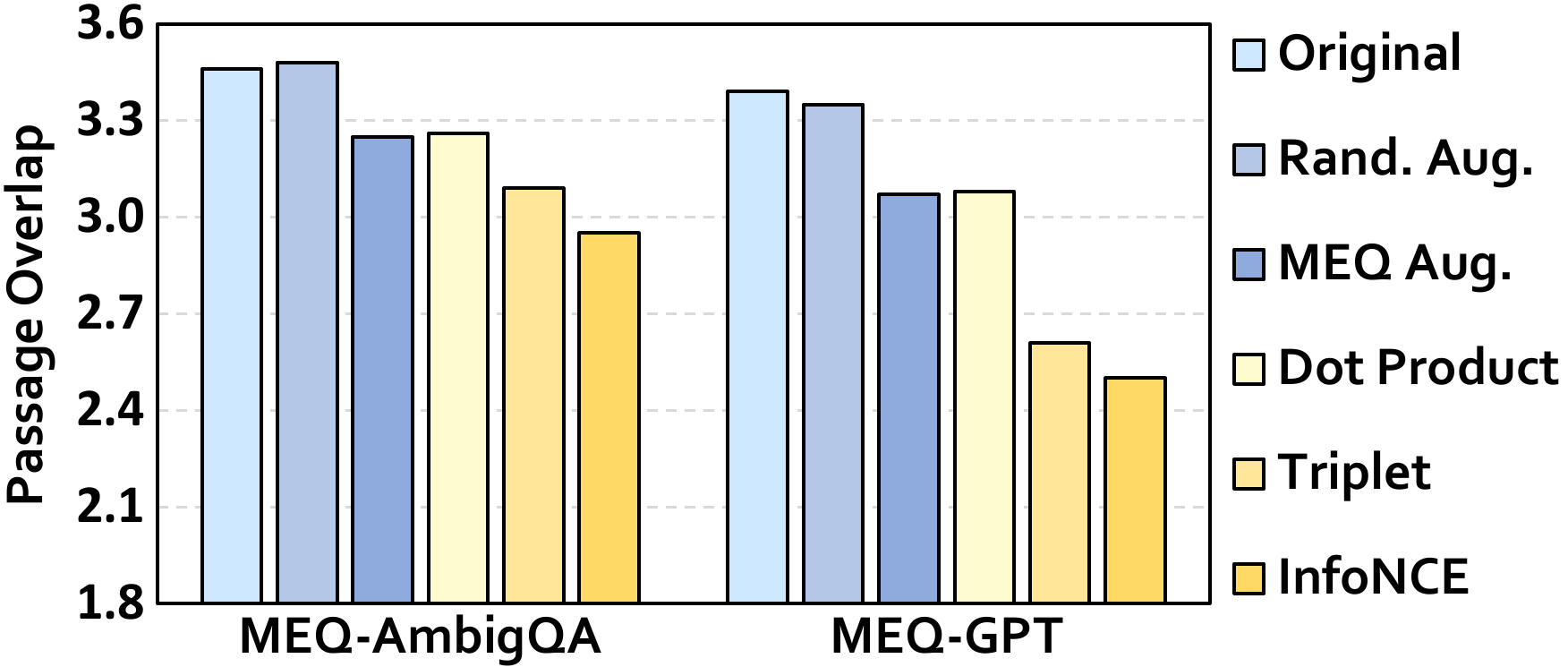}
    \caption{Overlap in top-5 retrieved passages between the original training question and its MEQ.}
    \label{fig:overlap}
\end{figure}
\subsection{Analysis}
\paragraph{Passage overlap}~{
One of the indications that DPR lacks the ability to distinguish the original question and its MEQ is the high overlap between the passages retrieved for each. Figure~\ref{fig:overlap} illustrates that both synthetic data augmentation and the query-side contrastive loss can reduce passage overlap. The synthetic MEQ augmentation helps to train the question embeddings of MEQs closer to their positive passages. Moreover, the query-side contrastive loss explicitly trains the model to distinguish the original question and its MEQ apart. Nevertheless, a lower passage overlap does not always indicate better performance. For instance, our model with the dot product loss does not have the lowest passage overlap, but performs the best in retrieval evaluation.
}

\begin{figure}[t]
    \centering
    \includegraphics[width=0.44\textwidth]{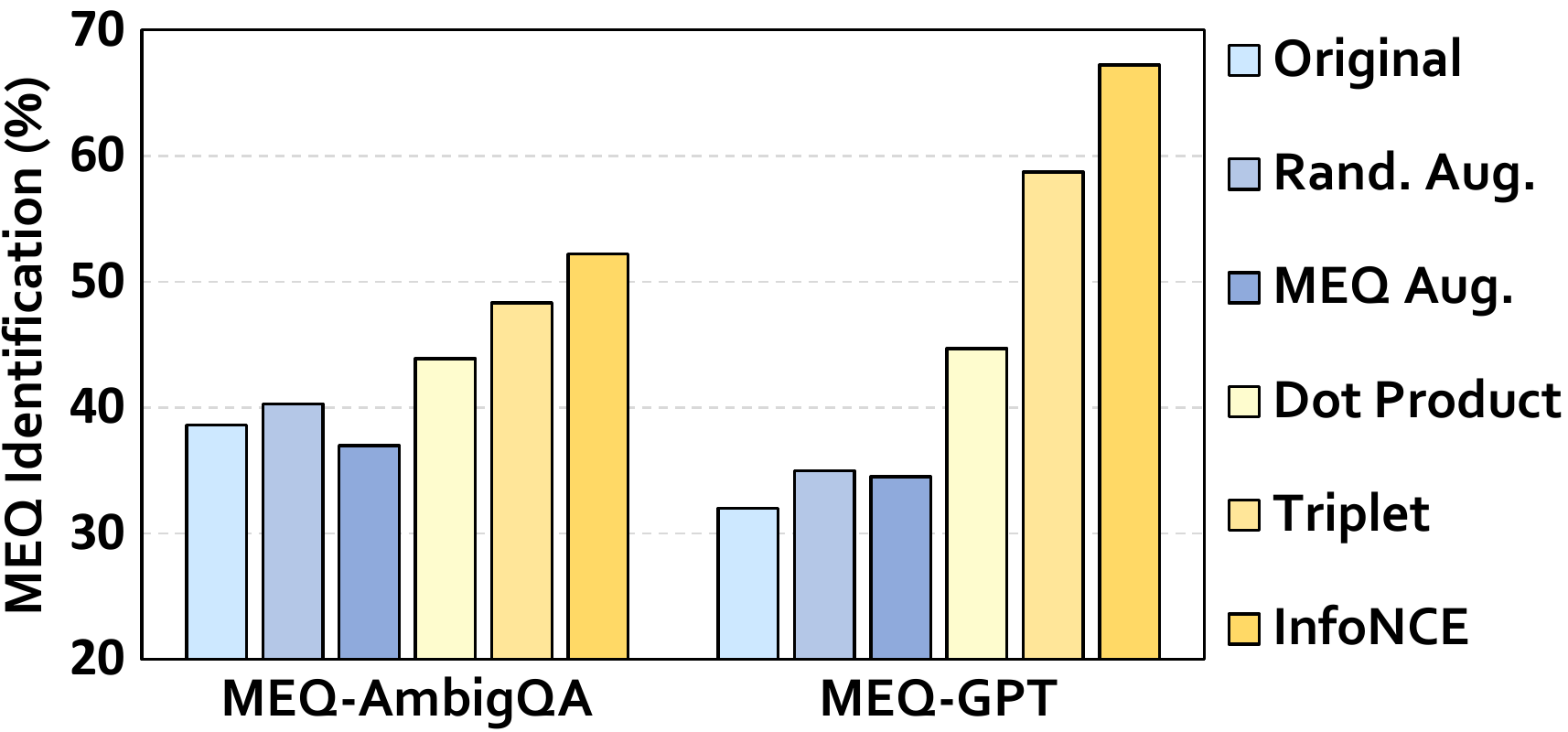}
    \caption{The ratio of successful MEQ identifications of different models on contrast sets, with paraphrase questions as distractors.}
    \label{fig:question_ranking}
\end{figure}

\paragraph{Identification of inter-question relationships}~{
To further analyze model behavior after the query-side contrastive training, we test the models' ability to distinguish inter-question relationships. A model is considered successful in identifying the MEQ if the generated embedding of the original question is closer to its paraphrase question rather than its MEQ. The paraphrase questions are separately generated using InstructGPT to avoid conflict with those used in data augmentation. As shown in Figure~\ref{fig:question_ranking}, training with the query-side contrastive loss leads to an improved ability to distinguish between paraphrase questions and different questions, which indicates our models are better at identifying inter-question relationships. The model trained with InfoNCE loss has the highest success rate in identifying inter-question relationships, because it received more training signals from a positive example and a set of negative examples than those with other types of loss.
}

\section{Conclusion}
In this study, we addressed the gap in research on contrast consistency in OpenQA by collecting MEQs as challenging contrast sets to the popular NQ benchmark. Our findings reveal that DPR lacks contrast consistency on our contrast sets. To address this limitation, we introduced a query-side contrastive loss with the aid of data augmentation, which improved its ability to recognize inter-question relationships. Overall, our findings and data can pave the way for further exploring the role of contrast consistency in developing robust and effective OpenQA systems.

% \newpage

% \section{Limitations}
% \input{8-Limitation.tex}

% \section*{Ethics Statement}
% Scientific work published at ACL 2023 must comply with the ACL Ethics Policy.\footnote{\url{https://www.aclweb.org/portal/content/acl-code-ethics}} We encourage all authors to include an explicit ethics statement on the broader impact of the work, or other ethical considerations after the conclusion but before the references. The ethics statement will not count toward the page limit (8 pages for long, 4 pages for short papers).

\section*{Acknowledgements}
This work was supported in part by NSF IIS-2119531, IIS-2137396, IIS-2142827, CCF-1901059, and ONR N00014-22-1-2507. Wenhao Yu is also supported in part by Bloomberg Data Science Ph.D Fellowship.
% This document has been adapted by Jordan Boyd-Graber, Naoaki Okazaki, Anna Rogers from the style files used for earlier ACL, EMNLP and NAACL proceedings, including those for

\bibliography{ref}
\bibliographystyle{acl_natbib}
  
\end{document}